\documentclass{article}

\usepackage{arxiv}
\usepackage{amsmath}
\usepackage{amssymb}
\usepackage{mathtools}
\usepackage{amsthm}
\usepackage[utf8]{inputenc} % allow utf-8 input
\usepackage[T1]{fontenc}    % use 8-bit T1 fonts
\usepackage{hyperref}       % hyperlinks
\usepackage{url}            % simple URL typesetting
\usepackage{booktabs}       % professional-quality tables
\usepackage{amsfonts}       % blackboard math symbols
\usepackage{nicefrac}       % compact symbols for 1/2, etc.
\usepackage{microtype}      % microtypography
\usepackage{lipsum}
\usepackage{graphicx}
\graphicspath{ {./images/} }
\usepackage{algorithm}
\usepackage{algorithmic}

\title{Thinking Like an Expert:Multimodal Hypergraph-of-Thought (HoT) Reasoning to boost Foundation Modals}

    \author{
    \textbf{Fanglong~Yao}$^{1,3,\dag}$, \textbf{Changyuan~Tian}$^{1,2,3,\dag}$, \textbf{Jintao~Liu}$^{1,2,3,\dag}$, \textbf{Zequn~Zhang}$^{1,3,\ast}$, \\\textbf{Qing~Liu}$^{1,3}$, \textbf{Li~Jin}$^{1,3}$,
    \textbf{Shuchao~Li}$^{1,3}$, 
    \textbf{Xiaoyu~Li}$^{1,3}$,\textbf{Xian~Sun}$^{1,2,3}$\thanks{Corresponding author. $^\dag$ Equal contribution.}
 \\
 $^{1}$Aerospace Information Research Institute, Chinese Academy of Sciences, Beijing 100190, China\\
  $^{2}$School of Electronic, Electrical and Communication Engineering, University of Chinese Academy of Sciences, China\\
  $^{3}$Key Laboratory of Network Information System Technology (NIST), Aerospace Information Research Institute, \\
  Chinese Academy of Sciences, Beijing, China \\
\texttt{\{yaofanglong17,tianchangyuan21,liujintao201\}@mails.ucas.ac.cn},\\
\texttt{zqzhang1@mail.ie.ac.cn,\{liuqing1,jinli,lisc,lixy01,sunxian\}@aircas.ac.cn
}
  \\
}

\begin{document}
\maketitle
\begin{abstract}
% This year is known as the ``Year of Foundation Models'' for hundreds of foundation models represented by chatGPT have been born.
Reasoning ability is one of the most crucial capabilities of a foundation model, signifying its capacity to address complex reasoning
tasks. Chain-of-Thought  (CoT) technique is widely regarded as one of the effective methods for enhancing the reasoning ability of foundation models and has garnered significant attention. However, the reasoning process of CoT is linear, step-by-step, similar to personal logical reasoning, suitable for solving general and slightly complicated problems. On the contrary, the thinking pattern of an expert owns two prominent characteristics that cannot be handled appropriately in CoT, i.e., high-order multi-hop reasoning and multimodal comparative judgement. Therefore, the core motivation of this paper is transcending CoT to construct a reasoning paradigm that can think like an expert. The hyperedge of a hypergraph could connect various vertices, making it naturally suitable for modelling high-order relationships. Inspired by this, this paper innovatively proposes a multimodal Hypergraph-of-Thought (HoT) reasoning paradigm, which enables the foundation models to possess the expert-level ability of high-order multi-hop reasoning and multimodal comparative judgement. Specifically, a textual hypergraph-of-thought is constructed utilizing triple as the primary thought to model higher-order relationships, and a hyperedge-of-thought is generated through multi-hop walking paths to achieve multi-hop inference. Furthermore, we devise a visual hypergraph-of-thought to interact with the textual hypergraph-of-thought via Cross-modal Co-Attention Graph Learning for multimodal comparative verification. Experimentations on the ScienceQA benchmark demonstrate the proposed HoT-based T5 outperforms CoT-based GPT3.5 and chatGPT, which is on par with CoT-based GPT4 with a lower model size.
\end{abstract}

% keywords can be removed

\section{Introduction}
% Since the release of ChatGPT~\cite{DBLP:journals/corr/abs-2304-09842} this year, the development of foundation models has exploded, with hundreds of fungible models emerging in competition~\cite{yang2023harnessing,chowdhery2022palm,zhang2022opt,workshop2023bloom,du2022glam,rae2022scaling,hoffmann2022training,thoppilan2022lamda,touvron2023llama,openai2023gpt4,wu2023bloomberggpt}. 
The immense success of ChatGPT ~\cite{DBLP:journals/corr/abs-2304-09842} has triggered an explosive growth in foundation models, resulting in the emergence of hundreds of alternative models competing in the field ~\cite{yang2023harnessing,chowdhery2022palm,zhang2022opt,workshop2023bloom,du2022glam,rae2022scaling,hoffmann2022training,thoppilan2022lamda,touvron2023llama,openai2023gpt4,wu2023bloomberggpt}. 
Reasoning ability is one of the most crucial capabilities of a foundation model, signifying its capacity to address complex reasoning tasks, such as science question answering \cite{lu2022learn}. Currently, one of the most effective techniques for enhancing the reasoning ability of foundation models is the Chain-of-Thought (CoT) \cite{wei2023chainofthought}. The key idea behind this approach is to guide the foundation model to generate additional intermediate reasoning steps, rather than just providing the final answer.
% The explosive increase in foundation models is mainly attributed to the powerful person-like reasoning ability endowed by the Chain-of-Thought (CoT)~\cite{brown2020language}, which is the phenomenon of ``intelligent emergence". Taking QA as an example, unlike the traditional approach of directly providing answers based on given questions~\cite{lan2021survey,10.1145/3289600.3290956}, the foundation model fine-tuned via CoT can provide the intermediate process of answering questions~\cite{wei2023chainofthought}. 
Building upon this key idea, researchers have successively introduced methods like CoT-SC~\cite{wang2023selfconsistency}, Skeleton-of-Thought (SoT)~\cite{ning2023skeletonofthought},Tree-of-Thought (ToT)~\cite{yao2023tree}, Graph-of-Thought (GoT)~\cite{yao2023chainofthought}.

Although the CoT endows foundation models with the ability to think like a person, the following limitations remain. Firstly, the CoT is linear, making it challenging to achieve leaping logical reasoning. Secondly, CoT decomposes a problem into step-by-step sequential procedures, with inadequate consideration for multi-concurrent steps and conflicts between them. The limitations restrict foundation models to solving non-professional problems.
 \begin{figure}
	 \centering
			\includegraphics[width=1\linewidth]{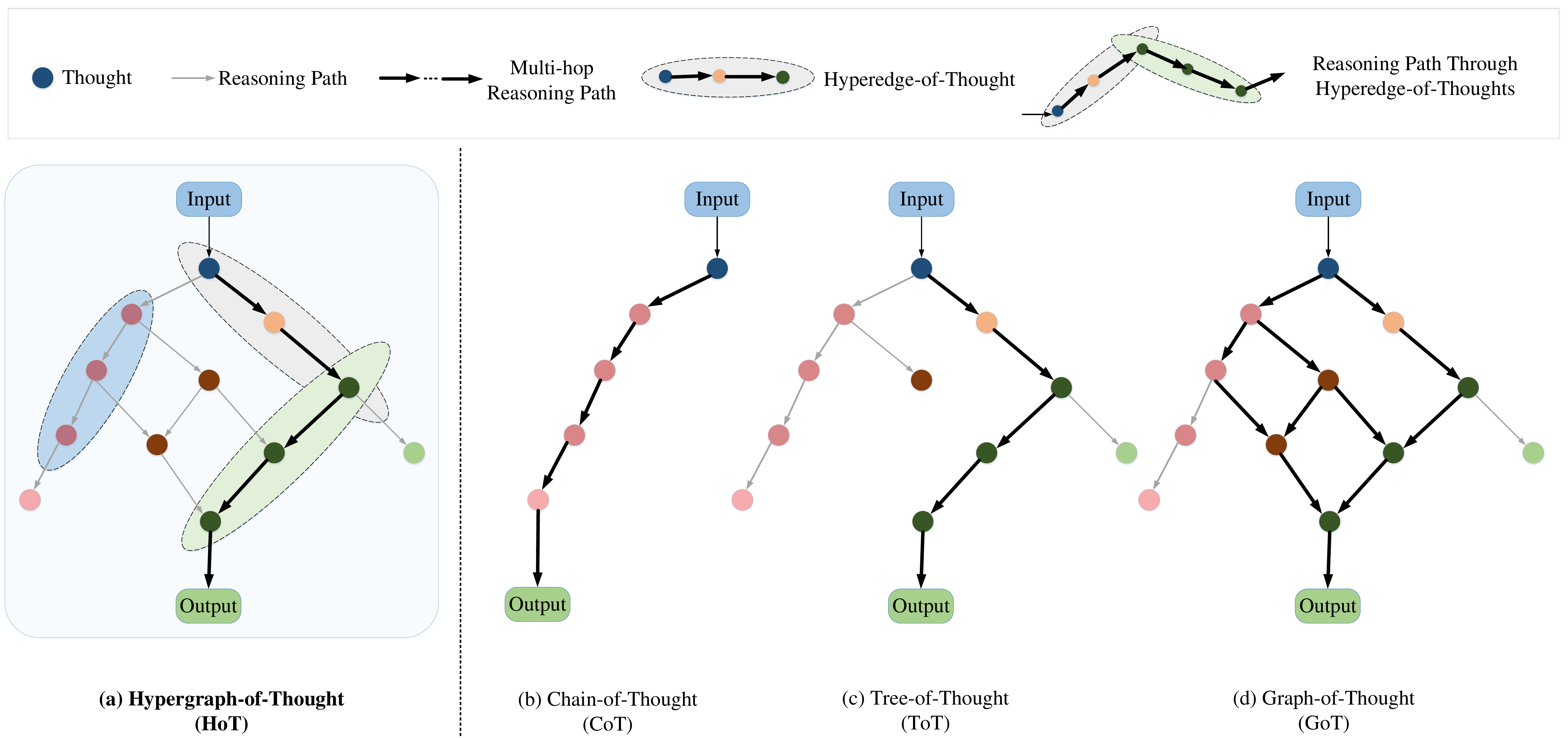}
\caption{ The paradigms of Hypergraph-of-Thought (HoT),  Chain-of-Thought (CoT), Tree-of-Thought (ToT), and Graph-of-Thought (GoT). Using HoT as the reference, removing the hyperedge-of-thought and retaining only one reasoning path becomes the CoT; Preserving multiple non-intersecting parallel reasoning paths constitutes the ToT; Simplifying the hyperedge-of-thought to a regular edge that can only connect two thoughts forms the GoT. }
\label{HoT_intro}
	\end{figure}\textbf{}

Compared to an ordinary person, the logical thinking mode of experts has two prominent characteristics,  i.e., high-order multi-hop reasoning and multimodal comparative judgement. Therefore, the core motivation of this paper is to concentrate on simulating the logical thinking of experts and constructing a reasoning paradigm that goes beyond the CoT to enhance the ability of foundation models to solve complex professional problems. Hypergraph is a powerful modelling approach for complex networks that has been widely studied recently. Compared with regular graphs that can only connect two vertices with con an edge, the hyperedge of hypergraph can connect an arbitrary number of vertices, so it is naturally suitable for modelling high-order relationships~\cite{10068184}. Compared to previously modelling the logical thinking process of ordinary people as a chain structure of sequences, we further simulate the logical thinking of experts directly and formulate it as a hypergraph structure, i.e., Hypergraph-of-Thought (HoT). Meanwhile, as illustrated in Figure~\ref{HoT_intro}, we believe HoT can be appropriately adjusted and degenerated into CoT, ToT, and GoT\footnote{It is worth noting that CoT-SC and SoT can be catgorized within the ToT reasoning paradigm.}.

This paper proposes an innovative multimodal hypergraph-of-thought reasoning paradigm, which empowers the foundation model with expert-like high-order reasoning capability based on the representation ability of textual and visual foundation models:
\begin{itemize}
\item Constructing a textual hypergraph-of-thought by utilizing triplets as basic thoughts to model higher-order relationships. Exploiting multi-hop random walks to form long-range reasoning paths to constitute hyperedges-of-thoughts and form multi-hop reasoning abilities.

\item Simultaneously, k-means clusters image patches to construct hyperedges to form a visual hypergraph-of-thought.

\item Allset Transformer is selected to encode the thoughts of the textual and visual HoTs, respectively, to achieve bidirectional updates of thought-to-hyperedge and hyperedge-to-thought.
\end{itemize}

In addition, to simulate the expert's multimodal contrastive judgement capability, we pick out Cross-modal Co-Attention Graph Learning to fully interact between different HoTs to avoid information conflicts from various modalities. We choose ScienceQA as the experimental dataset and follow the two-stage framework of Multimodal CoT~\cite{DBLP:journals/corr/abs-2302-00923}, namely the rational and answer generation stages. The experimental results are superior to CoT-based GPT-3.5 and ChatGPT, demonstrating the potential of HoT to enhance the ability of foundation models.

The contributions can be summarized as follows:

(1) Inspired by the logical reasoning process of an expert to solve complex problems, the multimodal hypergraph-of-thought reasoning paradigm for text and image is innovatively constructed, boosting the foundation models possessing the abilities of high-order multi-hop reasoning and multimodal comparative judgement.

(2)We take triplet as the basic thought to construct a textual hypergraph-of-thought to model higher-order relationships, and devise hyperedge-of-thought through multi-hop walking paths to achieve multi-hop inference. Further, utilizing AllSet transformer to achieve alternating updates of thought-to-hyperedge and hyperedge-to-thought.

(3)This paper also constructs an visual hypergraph-of-thought and utilizes Cross-modal Co-Attention Graph Learning to interact with the textual hypergraph-of-thought, achieving multimodal interactive verification.

\section{Related work}
\subsection{CoT}

The core idea of the COT \cite{wei2023chainofthought} is to guide LLMs in the process of reasoning by generating a series of intermediate inference steps, rather than directly providing the final answer. This reasoning mechanism effectively enhances the complex reasoning ability of LLMs. Researchers have further developed CoT from various perspectives, building upon this fundamental idea. In terms of paradigms, these efforts can be categorized as prompt-based CoT and fine-tuning-based CoT.  As for the targeted modalities, these endeavours can be classified as language-modal CoT and multi-modal CoT.

In the current research landscape, prompt-based CoT represents the mainstream approach. Based on whether a few demonstrations are provided to the LLMs, prompt-based CoT techniques can be categorized into few-shot CoT \cite{wei2023chainofthought} and zero-shot CoT\cite{kojima2023large}. Typically, standard CoT prompting falls under the category of few-shot CoT prompting, where specific demonstrations must be provided as examples to guide LLMs' reasoning process \cite{wei2023chainofthought}. On the other hand, zero-shot CoT requires no additional demonstrations; instead, adding the prompt "Let's think step by step" between each answer can enhance the complex reasoning ability of LLMs.

Although prompt-based CoT techniques require no parameter adjustments, one major drawback is their reliance on extensive models with billions of parameters, making them challenging to deploy on a large scale in practical applications. Researchers have been investigating methods to imbue smaller models with similar reasoning capabilities to address this issue and reduce the model size requirements for CoT. Fine-tuning-based CoT is considered a promising solution in this regard. Fine-tuning-based CoT involves fine-tuning a smaller model (with fewer than 1 billion parameters) to generate informative rationales, which are then used for reasoning and producing the final answers, rather than providing direct responses. Namgyu Ho et al. \cite{ho2022large} proposed fine-tune-CoT, which leverages prompt-based CoT with a large teacher model to generate reasoning samples. These samples are then used for fine-tuning a smaller model, endowing it with significant reasoning abilities.
Similarly, Lucie Charlotte Magister et al. \cite{magister2022teaching} explored knowledge extraction from large language models like PaLM 540B and GPT-3 175B to smaller models with different sizes, such as T5 XXL, XL, and base, which have parameters of 11 billion, 3 billion, and 220 million, respectively. With similar motivation, Liunian Harold Li et al. \cite{li2023symbolic} introduced Symbolic CoT Distillation, which samples rationales from a large teacher model to train a smaller student model. These efforts demonstrate promising advances in enabling smaller models to exhibit effective reasoning capabilities in CoT, allowing for more practical and scalable deployments in real-world scenarios.

The capabilities of CoT should not be limited to language models alone. As a result, researchers have been striving to expand CoT beyond the language modality into the realm of multi-modal tasks, giving rise to Multi-modal CoT. Building upon the core idea of CoT, Multi-modal CoT aims to facilitate multi-step reasoning for multi-modal inference tasks, rather than providing direct answers. Zhang et al. \cite{zhang2023multimodal} introduced the MM-CoT framework, representing the first attempt to extend the CoT reasoning mechanism to multi-modal scientific question-answering tasks. This extension significantly improves reasoning performance. Specifically, MM-CoT is a two-stage framework. In the first stage, it generates rationales, and in the second stage, based on the rationales generated in the previous stage, it carries out the final answer inference. Expanding on the MM-CoT framework, Yao et al. \cite{yao2023beyond} proposed an enhancement by incorporating additional graphs to model human thinking processes. Their approach, called GoT, leverages the graph modality to improve reasoning performance.

\subsection{Vision-Language LLMs}
Recently, there has been significant interest in fine-tuning LLMs for vision-language instruction following. Researchers have primarily explored methods for fine-tuning LLMs using multi-modal instruction-following data. Liu et al. \cite{liu2023visual} propose groundbreaking research on generating multi-modal language-image instruction data using GPT-4, which only contains the language modality. Through tuning on the generated data, they introduce LLaVA, a large multi-modal model that connects visual encoders and LLMs for general visual and language understanding.
Similarly, Luo et al. \cite{luo2023cheap} present an efficient and economical approach for fine-tuning LLMs in the context of vision-language instructions. They introduce the MMA (Mixture-of-Modality Adaptation) framework, utilizing lightweight adapters to connect LLMs with vision-language tasks. Wang et al. \cite{wang2023t} address the challenge of gathering high-quality CoT rationales for answering scientific questions in multi-modal scenarios. They propose the T-SciQ method, which generates high-quality CoT rationales as teaching signals and trains smaller models for CoT reasoning. Horawalavithana et al. \cite{horawalavithana2023scitune} introduce a tuning framework called SciTune to align LLMs with scientific multi-modal instructions. They emphasize the limited research on improving LLMs to align with scientific subjects, concepts, and objectives. By fine-tuning LLMs using human-generated scientific instruction data, they train the LLaMA-SciTune, a large multi-modal model connecting visual encoders and LLMs.

\section{Method}

\begin{figure*}
    \centering
    \includegraphics[width=1\textwidth]{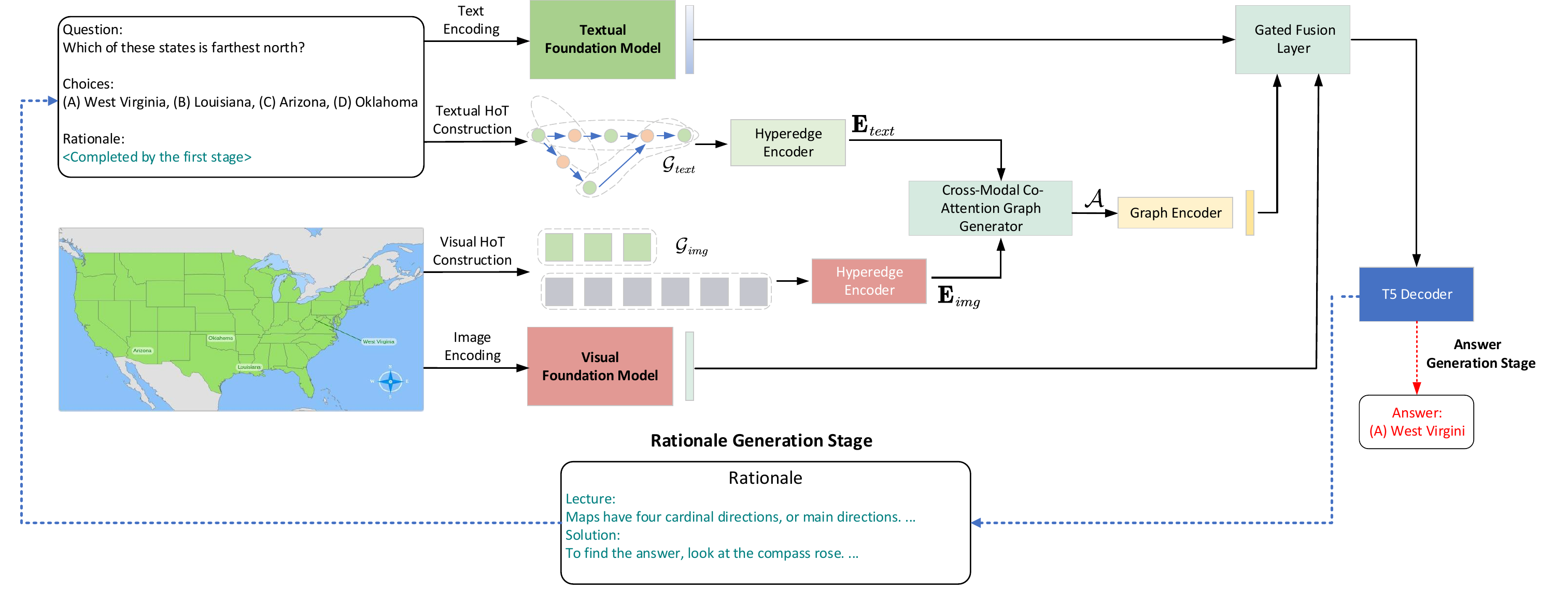}
    \caption{Similar to \cite{DBLP:journals/corr/abs-2302-00923} \cite{yao2023chainofthought}, we employ a two-stage framework for implementing CoT reasoning, comprising the Rationale Generation and Answer Generation stages. The input in the Rationale Generation stage consists of textual (question) and image modalities. Initially, we employ separate text and image encoders to encode textual and image information, respectively. The parameters of textual and visual foundation models are frozen during the training process. Additionally, we construct textual and visual HoTs for the respective modalities , and utilize the AllSetTransformer hypergraph encoder to encode these hypergraphs. Subsequently, we establish a cross-modal cross-attention graph, obtaining a cross-modal representation through the graph encoder. Following the gate-based fusion layer, the fused representation of textual, visual, and cross-modal information is fed into the decoder to generate intermediate rationales. In the Answer Generation stage, the distinction from the previous stage lies in both input and output. This stage's input encompasses the previous stage's output, i.e., the rationale, and includes additional contextual information. The output of this stage culminates in the final question answer.}
    \label{fig:enter-label}
\end{figure*}

\renewcommand{\algorithmicrequire}{\textbf{Input:}}
\renewcommand{\algorithmicensure}{\textbf{Output:}}
\begin{algorithm}
    \caption{Textual Hypergraph-of-Thought Construction}
    \begin{algorithmic}
        \ENSURE{$\mathcal{G}_{text}(V_{text}, \mathcal{E}_{text})$}
        \REQUIRE{$\mathcal{T}$, $G(V, E)$, k, N}
        \STATE $\mathcal{E}_{text}$ is empty
        \FOR{n=1 to N} 
        \STATE \quad $v_{start}$ $\longleftarrow$ RandomChoice($V$)
        \STATE \quad $e_{text}$ $\longleftarrow$ RandomWalk($v_{start}$, $G$, k)
        \STATE Append $e_{text}$ to $\mathcal{E}_{text}$
        \ENDFOR
        \STATE Return $\mathcal{G}_{text}(\mathcal{V}_{text}, \mathcal{E}_{text})$
    \end{algorithmic}
\end{algorithm}

\begin{algorithm}
    \caption{Visual Hypergraph-of-Thought Construction}
    \begin{algorithmic}
        \ENSURE{$\mathcal{G}_{img}(V_{img}, \mathcal{E}_{img})$}
        \REQUIRE{$\mathcal{I}$, m}
        \STATE $\mathbf{P} \longleftarrow Swin(\mathcal{I})$
        \STATE $\mathcal{G}_{img} \longleftarrow$k-means($\mathbf{P}$, m)
        
        \STATE Return $\mathcal{G}_{img}$
    \end{algorithmic}
\end{algorithm}

\subsection{Hypergraphof-Thought Construction}
Hypergraphs offer a natural way to model higher-order relationships, where each hyperedge can connect any number of objects. Previous research has demonstrated the application of hypergraphs in modelling higher-order semantics and improving capabilities in multi-hop reasoning and effective multi-modal learning\cite{heo2022hypergraph},\cite{kim2020hypergraph}. Based on this observation, utilizing hypergraphs to model essential higher-order relationships and aligning semantic information from different modalities can enhance the model's reasoning abilities. Therefore, our first step involves constructing hypergraphs for the text and image modalities separately, aiming to capture the higher-order semantic relationships within each modality. Additionally, inspired by the work of \cite{kim2020hypergraph}, we further construct a multi-modal hypergraph to align the semantic information between the image and text modalities.

\subsubsection{Textual Hypergraph-of-Thought Construction}

Previous research has empirically demonstrated that using thought graphs to model human's ability for leaps of reasoning can improve the reasoning capabilities of vision-language LLMs. Taking this a step further, and inspired by \cite{heo2022hypergraph}, we believe that using hypergraphs to model the human ability for leaps of reasoning is more appropriate. Hypergraphs can capture higher-order reasoning paths and enhance multi-hop reasoning capabilities.

\begin{figure}
    \centering
    \includegraphics[width=1\textwidth]{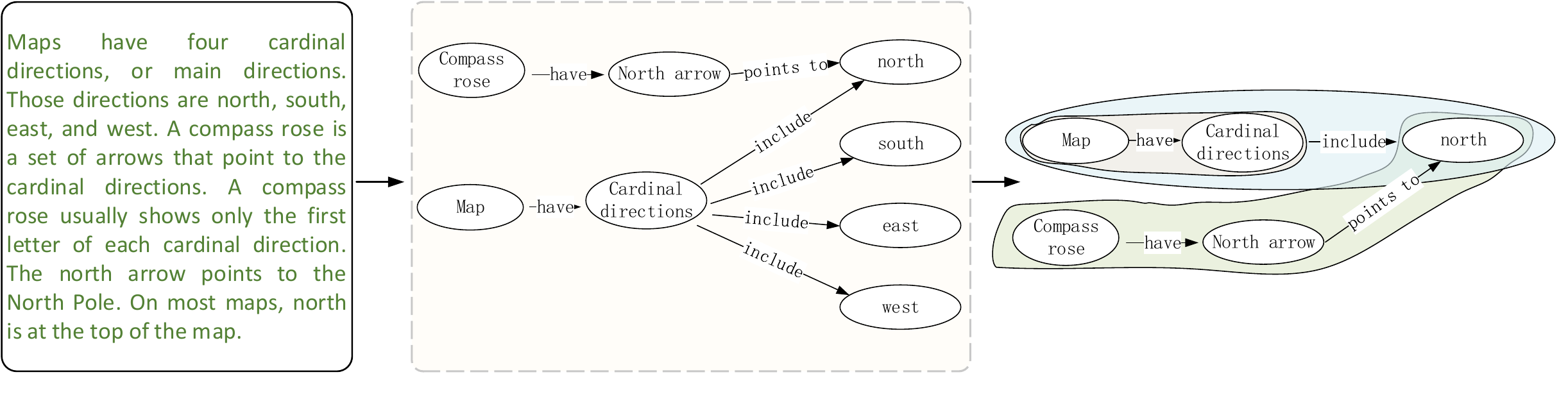}
    \caption{Example of Textual Hypergraph-of-Thought Construction}
    \label{fig:enter-texthyper}
\end{figure}

As shown in figure~\ref{fig:enter-texthyper}, following \cite{yao2023beyond}, we first construct a graph-of-thoughts, denoted as $G=(V, E)$, where $V$ represents the set of thoughts and $E$ represents the connections between thoughts. In the text hypergraph $\mathcal{G}_{text}=(\mathcal{V}_{text}, \mathcal{E}_{text})$, the text hypergraph shares the same node set as the graph of thoughts, i.e., $V=\mathcal{V}_{text}$. Hyperedge $e_{text} \in \mathcal{E}_{text}$ is defined as a multi-hop reasoning path, such as (Lionel Messi, place of birth, Rosario, is located in, Republic of Argentina, is located in, South America). Multi-hop reasoning paths are obtained through random walks on the graph of thoughts. It's worth noting that we define a triplet as the basic unit of random walks. We have set two types of random walks: one-hop random walks and k-hop random walks. One-hop random walks correspond to triple hyperedges, such as (Lionel Messi, place of birth, Rosario). On the other hand, k-hop random walks refer to k consecutive walks, meaning that the starting point of the current walk is the endpoint of the last walk. An example of a 2-hop walk is (Lionel Messi, place of birth, Rosario, is located in, Republic of Argentina).

\subsubsection{Visual Hypergraph-of-Thought Construction}

We believe that hypergraphs can capture high-order interactions among various local objects in images, enhancing the modelling capability of global image information\cite{TIAN2022263}. Therefore, we additionally construct an image hypergraph. 

Firstly, we utilize Swin\cite{liu2021swin} to perform representation learning on the initial input images, obtaining representations for a set of patches denoted as $\mathbf{P} \in \mathbf{R}^{p \times d}$, where $p$ represents the number of patches obtained by dividing the input image. Let $\mathcal{I}$ denote the image, then we have:

   $$ \mathbf{P} = \text{Swin}(\mathcal{I})$$

Inspired by the hypergraph construction method introduced in \cite{gaohyper}, we use the K-means clustering algorithm to generate the image hypergraph $\mathcal{G}_{img}=(\mathcal{V}_{img}, \mathcal{E}_{img})$. Specifically, each cluster is considered a hyperedge, so $|\mathcal{E}_{img}|=m$, where $m$ is a hyperparameter representing the number of clusters. We denote this process as:

$$
    \mathcal{G}_{img} = \text{k-means}(m, \mathbf{P})
$$

\subsection{Textual and Visual Hypergraph-of-Thought Encoding}

\subsubsection{Hypergraph Encoder}

Firstly, we define the hypergraph encoder. Many hypergraph neural networks have been proposed to learn hypergraph representations effectively. In our work, we use the AllSet Transformer as the hypergraph encoder. The AllSet Transformer consists of two multisets functions, namely the node-to-hyperedge function: $f_{\mathcal{V} \rightarrow \mathcal{E}}$, and the hyperedge-to-node function: $f_{\mathcal{E} \rightarrow \mathcal{V}}$. Transformers parameterize both functions. Given a matrix $\mathbf{S} \in R^{|S|\times d}$, representing a d-dimensional vector for each element in the multiset $S$, the AllSet Transformer is defined as follows:

$$f_{\mathcal{V} \rightarrow \mathcal{E}}(S)=f_{\mathcal{E} \rightarrow \mathcal{V}}(S)=\text{LN}(\mathbf{Y}+\text{MLP}(\mathbf{Y}))$$

$$\mathbf{Y}=\text{LN}\left(\theta+\mathbf{M H}_{h, \sigma}(\theta, \mathbf{S}, \mathbf{S})\right), \mathbf{M H}_{h, \sigma}(\theta, \mathbf{S}, \mathbf{S})=\big\|_{i=1}^h \mathbf{O}^{(i)}$$

$$\mathbf{O}^{(i)}=\sigma\left(\theta^{(i)}\left(\mathbf{K}^{(i)}\right)^T\right) \mathbf{V}^{(i)}$$
$$\theta \triangleq \big\|_{i=1}^h \theta^{(i)}$$
$$\mathbf{K}^{(i)}=\text{MLP}^{K, i}(\mathbf{S})$$
$$\mathbf{V}^{(i)}=\text{MLP}^{V, i}(\mathbf{S})$$

where LN denotes layer normalization, $h$ represents the number of heads in the multi-head attention mechanism, $\theta \in R^{1 \times h d_{h}}$ are learnable weights. $||$ indicates the concatenation operation, and the output dimensions of $MLP^{K, i}$ and $MLP^{V, i}$ are $R^{|\mathcal{V}| \times d_h}$.

\subsubsection{Textual Hypergraph-of-Thought Encoding}

Since the initial nodes on the text hypergraph are expressed as thoughts represented by words, we first need to obtain embeddings for these thoughts. Following the setup of \cite{yao2023beyond}, we format $\mathcal{V}_{text}$ as a sequence, i.e., 

$$p = [<s>, n_0, </s>, ..., <s>, n_j, </s>]$$

where $<s>$ and $</s>$ are special tokens used to emphasize each graph of thought node, and $n_i$ represents a node in the graph of thought. $p$ represents the formatted token sequence. Then, we input the token sequence into the text encoder, and the representations of each node corresponding to $<s>$ output by the encoder are considered as the initial representations of the nodes, i.e.,

$$
\left[x_0^s, x_0^n, x_0^e, \ldots, x_{|\mathcal{V}|}^s, x_{|\mathcal{V}|}^n, x_{|\mathcal{V}|}^e\right]=\text{Encoder}_{text}(p)
$$

where $x_i^s$ is considered as the initial representation of node $n_i$. Thus, we have $\mathbf{X}=[x_0^s;...;x_{|\mathcal{V}|}^s]\in R^{{|\mathcal{V}|} \times d}$, where $d$ represents the dimension of the embeddings.

Next, with the text hypergraph $\mathcal{G}_{text}$ and the initial node embeddings $\mathbf{X}$ as inputs, we utilize the AllSet Transformer encoder to encode the hypergraph nodes.

\subsubsection{Visual Hypergraph-of-Thought Encoding}

Unlike the text hypergraph, the nodes on the image hypergraph do not require an additional embedding layer, as their initial representations $\mathbf{P} \in \mathbf{R}^{p \times d}$ are obtained from the Swin encoder. Similar to the encoding process for the text hypergraph, we use the AllSet Transformer to encode the image hypergraph.

\subsection{Cross-HoT Interaction}

The textual and visual HoTs model high-order semantic information from the text and image modalities. To enhance cross-modal interaction, we introduce the Cross-Modal Co-Attention Graph Learning module. The core idea of the co-attention graph is to enable interaction between the hyperedges $\mathcal{E}_{text}$ of the text hypergraph and $\mathcal{E}_{img}$ of the image hypergraph. First, we use the multisets function $f_{\mathcal{V}\rightarrow \mathcal{E}}$ introduced in 3.2.2 to learn the representations of the hyperedges, i.e.,

$$
\mathbf{e}_{text}^i= f_{\mathcal{V}_{text} \rightarrow \mathcal{E}_{text}}(\mathbf{S_{text}^i})
$$

$$
\mathbf{e}_{img}^i= f_{\mathcal{V}_{img} \rightarrow \mathcal{E}_{img}}(\mathbf{S_{img}^i})
$$

where $\mathbf{e}_{text}^i \in \mathcal{E}_{text}$ represents the representation of the i-th hyperedge in the text hypergraph, and $\mathbf{S}_{text}^i \in R^{|S_{text}^i| \times d}$ is the set of vectors representing the i-th hyperedge in the text hypergraph, with each row representing the representation of a node. Similarly, $\mathbf{e}_{img}^i \in \mathcal{E}_{img}$ represents the representation of the i-th hyperedge in the image hypergraph, and $\mathbf{S}_{img}^i \in R^{|S_{img}^i| \times d}$ is the set of vectors representing the i-th hyperedge in the image hypergraph, with each row representing the representation of a node (patch). Thus, we have the text hyperedge representation $\mathbf{E}_{text} \in R^{|\mathcal{E}_{text}| \times d}$ and the image hyperedge representation $\mathbf{E}_{img} \in R^{|\mathcal{E}_{img}| \times d}$.

Next, based on the text and image hyperedges representations, we use vector inner product to construct the Cross-Modal Co-Attention Graph. This process is represented as follows:

$$
\mathcal{A}=\text{softmax}\left(W \circ\left(E_{text} W_{text}^c \right)\left(E_{img} W_{img}^c \right)^{\top}\right)
$$

where $W_{text}^c \in R^{d \times d_c}$, $ W_{img}^c \in R^{d \times d_c}$, and $W \in R^{|\mathcal{E}_{text}| \times |\mathcal{E}_{img}|}$ are learnable weights.

$$
\mathbf{z}_{m}= \left(E_{text} W_{text}^m\right)^{\top} \mathcal{A}\left(E_{img} W_{img}^m\right)
$$

where $\mathbf{z}_m$ represents the final fused representation, and $\mathbf{W}_{text}^m \in R^{d \times d_m}$ and $\mathbf{W}_{img}^m \in R^{d \times d_m}$ are learnable weights.

\section{Experiments}
\subsection{Dataset}
Our method has been assessed using the ScienceQA benchmark \cite{lu2022learn} is a pioneering multimodal science question dataset. Notably, ScienceQA provides detailed lectures and explanations annotated alongside the answers. The dataset comprises 21000 multiple-choice questions, covering various domains across three subjects, 26 topics, 127 categories, and 379 skills. To facilitate evaluation, the benchmark dataset is divided into training, validation, and test splits, consisting of 12726, 4241, and 4241 examples, respectively.

\subsection{Implementation Details}
Our experiments are conducted on 6 NVIDIA Tesla V100 32G GPUs. In our experimental setup, we employed the T5 architecture \cite{raffel2020exploring} as our baseline model, utilizing both T5-base and T5-large model sizes. We initialized our model using the pre-trained T5 checkpoint, specifically the UnifiedQA \cite{DBLP:conf/emnlp/KhashabiMKSTCH20} variant to ensure a fair comparison. We adopt DETR \cite{DBLP:conf/eccv/CarionMSUKZ20} for the visual foundation modal to obtain visual features. Our models underwent fine-tuning for 20 epochs, employing a learning rate 5e-5. The maximum input sequence length is set to 512. We set the batch sizes for the base and large models to 8 and 4, respectively.

\subsection{Baselines}
Following previous methods \cite{lu2022learn,DBLP:journals/corr/abs-2302-00923}, we select the following baselines for comparison: 

\begin{itemize}

    \item  Visual question answering (VQA) models \cite{DBLP:conf/cvpr/Yu0CT019,DBLP:conf/cvpr/00010BT0GZ18,DBLP:conf/nips/KimJZ18,DBLP:conf/cvpr/GaoJYLHWL19,DBLP:conf/nips/LuQCXZZYLZ21,DBLP:journals/corr/abs-1908-03557,DBLP:conf/icml/KimSK21}. These methods treat the question, context, and choices as the textual input and the
image as the vision input. Then they predict the score distribution over choice candidates via a linear classifier.

    \item  Text-to-text LM models \cite{raffel2020exploring,DBLP:conf/nips/ChenKSNH20}. UnifiedQA \cite{raffel2020exploring} and GPT-3.5 \cite{DBLP:conf/nips/ChenKSNH20} regard this task as a text generation problem and the model is trained to generate the target text. Then they obtain the prediction results from the generated text. 

    \item  Text-to-text LLMs with CoT prompting \cite{lu2022learn}. It is worth noting that UnifiedQA and GPT-3.5 \cite{lu2022learn} adopt the generated image captions as vision semantics. 
\end{itemize}

\subsection{Main Results}

\begin{table*}[!t]
	\centering
	
	\renewcommand\arraystretch{1.2}
	\setlength\tabcolsep{8pt}
	
	\resizebox{15cm}{!}{
	\begin{tabular}{l| c| c ccccccc|c}
		%\caption{Table captions should be placed above the tables.}\label{tab1}
		
		\toprule[1pt]
		Methods&Size&NAT&SOC&LAN&TXT&IMG&NO&G1-6&G7-12&Avg \\
		\midrule
		Human &- &90.23& 84.97& 87.48& 89.60& 87.50& 88.10& 91.59& 82.42& 88.40\\
		\midrule
		MCAN\cite{DBLP:conf/cvpr/Yu0CT019}& 95M& 56.08& 46.23& 58.09& 59.43& 51.17& 55.40& 51.65& 59.72& 54.54\\
		Top-Down\cite{DBLP:conf/cvpr/00010BT0GZ18}& 70M& 59.50& 54.33& 61.82& 62.90& 54.88& 59.79& 57.27& 62.16& 59.02\\
		BAN\cite{DBLP:conf/nips/KimJZ18}& 112M& 60.88& 46.57& 66.64& 62.61& 52.60& 65.51& 56.83& 63.94& 59.37\\
		DFAF\cite{DBLP:conf/cvpr/GaoJYLHWL19}& 74M& 64.03& 48.82& 63.55& 65.88& 54.49& 64.11& 57.12& 67.17& 60.72\\
		ViLT\cite{DBLP:conf/icml/KimSK21}& 113M& 60.48& 63.89& 60.27& 63.20& 61.38& 57.00& 60.72& 61.90& 61.14\\
		Patch-TRM\cite{DBLP:conf/nips/LuQCXZZYLZ21}& 90M& 65.19& 46.79& 65.55& 66.96& 55.28& 64.95& 58.04& 67.50& 61.42\\
		VisualBERT\cite{DBLP:journals/corr/abs-1908-03557}& 111M& 59.33& 69.18& 61.18& 62.71& 62.17& 58.54& 62.96& 59.92& 61.87\\
		\midrule
		UnifiedQA\footnotesize Base \normalsize\cite{raffel2020exploring}&223M& 68.16& 69.18& 74.91& 63.78& 61.38& 77.84& 72.98& 65.00& 70.12\\
		GPT-3.5\cite{DBLP:conf/nips/ChenKSNH20}& 175B& 74.64& 69.74& 76.00& 74.44& 67.28& 77.42& 76.80& 68.89& 73.97\\
		\midrule
		UnifiedQA\footnotesize Base \normalsize (CoT)\cite{lu2022learn}&223M& 71.00& 76.04& 78.91& 66.42& 66.53& 81.81& 77.06& 68.82& 74.11\\
		GPT-3.5 (CoT)\cite{lu2022learn}& 175B &75.44& 70.87& 78.09& 74.68& 67.43& 79.93& 78.23& 69.68& 75.17\\
		ChatGPT (CoT)\cite{DBLP:journals/corr/abs-2304-09842}& - & 78.82& 70.98& 83.18& 77.37& 67.92& 86.13 &80.72& 74.03& 78.31\\
		GPT-4 (CoT)\cite{DBLP:journals/corr/abs-2304-09842}&  -& 85.48& 72.44& 90.27& 82.65& 71.49& 92.89& 86.66& 79.04& 83.99\\
		\midrule
        HoT-T5\footnotesize Base \normalsize & 223M &82.46&	78.07&	82.00&	81.18&	75.20&	85.09	&81.86&	80.62&	81.42\\
		HoT-T5\footnotesize Large \normalsize &738M&84.46&79.08&84.64&82.89&75.81&88.15&83.88&82.47&83.38\\
		\bottomrule[1pt]
	\end{tabular}}
	\caption{Overall performance compared to the baselines on the ScienceQA dataset.}
	\label{main}
\end{table*}

%\midrule%
		%Mutimodal-CoT\footnotesize Large \normalsize\cite{DBLP:journals/corr/abs-2302-00923}& 738M& 95.91& 82.00 &90.82& 95.26 &88.80& 92.89& 92.44& 90.31 &91.68\\%
		%HoT-T5\footnotesize Large \normalsize&738M&&&&&&&&&\\
  		%Mutimodal-CoT\footnotesize Base \normalsize\cite{DBLP:journals/corr/abs-2302-00923}& 223M& 87.52& 77.17& 85.82& 87.88& 82.90& 86.83& 84.65& 85.37& 84.91\\%
		%HoT-T5\footnotesize Base \normalsize&223M&&&&&&&&&\\
  
The main results are shown in Table \ref{main}. 
According to the evaluation results, we can observe that:

(1) HoT-T5 has demonstrated remarkable performance, outperforming previous methods by an impressive margin. Moreover, it has even surpassed human performance levels.
The performance improvement over baselines demonstrates that the proposed Textual HoT, Visual HoT, and Cross-HoT Interaction can improve the expert-level ability of high-order multi-hop reasoning and multimodal comparative judgement.

(2) Specifically, among the 8 question classes, HoT-T5\footnotesize Large \normalsize significantly improved for questions accompanied by paired images. This highlights the effectiveness of incorporating image features to enhance question-answering capabilities compared to methods that rely solely on image captions for vision semantics, such as UnifiedQA and GPT-3.5.

(3) The two-stage framework used in the HoT-T5\footnotesize Large \normalsize approach has also proven to contribute to its superior results, which validates the potential and effectiveness of leveraging multimodal inputs and image features for this task.
% improved question answering capabilities, making significant advancements in the field of natural language understanding and multimodal reasoning.

\subsection{Ablation Study}
\begin{table*}[!t]
	\centering
	
	\renewcommand\arraystretch{1.2}
	\setlength\tabcolsep{8pt}
	
	\resizebox{15cm}{!}{
		\begin{tabular}{l| c ccccccc|c}
			%\caption{Table captions should be placed above the tables.}\label{tab1}
			
			\toprule[1pt]
			Methods&NAT&SOC&LAN&TXT&IMG&NO&G1-6&G7-12&Avg \\
			\midrule
			w/o Textual HoT&79.97&	76.04&	80.64&	79.13	&72.58	&83.14&80.03&	78.05	&79.32\\
			w/o Visual HoT&80.95&77.39&81.82&80.01&74.22&84.18&80.98&79.43&80.43\\
			w/o Cross-HoT Interaction&81.22&77.84&82.82&79.96&74.42&85.37&81.28&80.29&80.92\\
			HoT-T5\footnotesize Base \normalsize&82.46&	78.07&	82.00&	81.18&	75.20&	85.09	&81.86&	80.62&	81.42\\
			
			\bottomrule[1pt]
			
	\end{tabular}}
	\caption{Experimental results on ablation studies.}
	\label{ablation}
\end{table*}

In this section, we conduct ablation studies to evaluate the contributions of each component. The experimental results are reported in Table \ref{ablation}. We can see that:

(1) After removing Textual HoT, the model performance drops significantly, demonstrating that the Textual HoT can provide rich semantic information about the text, thus improving the high-order multi-hop reasoning ability.

(2) After removing the Visual HoT, the model performance becomes worse. This is because the Visual HoT can help the model gain a deeper understanding of visual modalities and boost model performance.

(3) The model suffers from a performance decay after removing the Cross-HoT Interaction. The reason may be that the Cross-HoT Interaction can combine semantic information of text and images for multimodal comparative judgement. 

\subsection{Case Study}
\begin{figure}
	 \centering
			\includegraphics[width=15.5cm]{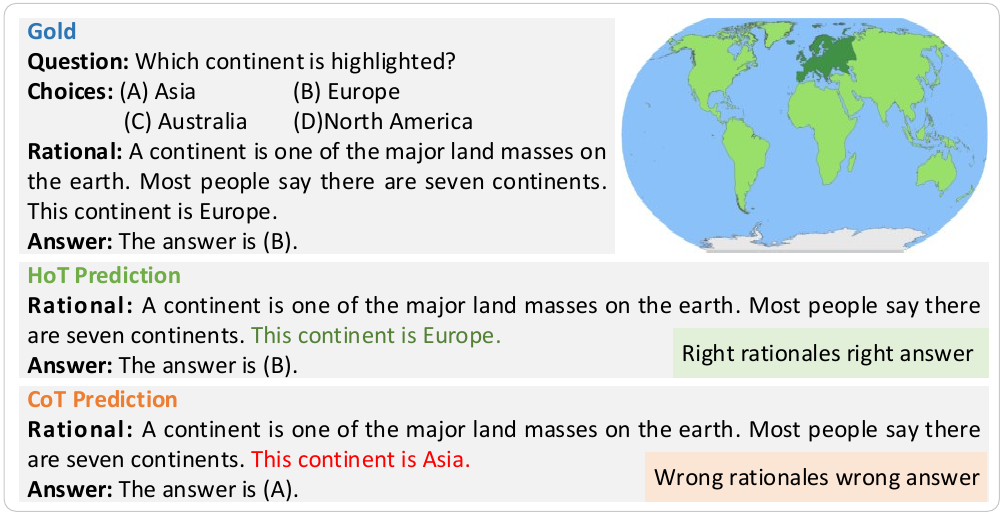}
\caption{Case study on the ScienceQA dataset.}
\label{case1}
	\end{figure}

 \begin{figure}
	 \centering
			\includegraphics[width=15.5cm]{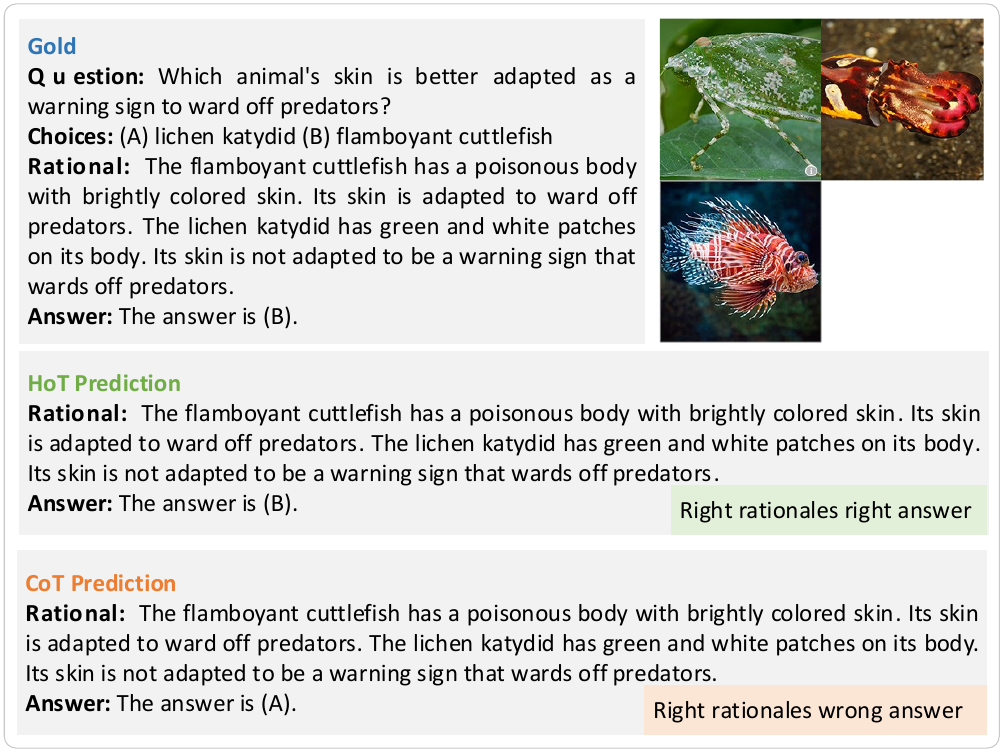}
\caption{Case study on the ScienceQA dataset. }
\label{case2}
	\end{figure}

To provide a more illustrative comparison between HoT and CoT, we have selected two representative examples from the ScienceQA dataset for analysis. In Figure \ref{case1}, we find that CoT produces wrong rations and answers, while HoT achieves the right results. This indicates that HoT can address high-order multi-hop reasoning and multimodal comparative judgement better. In Figure \ref{case2}, we can observe that CoT produces right rationals but wrong answers. HoT can leverage the generated rationales more effectively, thus achieving better performance than CoT.

\section{Conclusion}
This paper breaks through the previous chain-of-thought (CoT) pattern of being able to think like an ordinary person. Inspired by expert logical thinking and combined with the advantages of hypergraphs, we construct a multimodal hypergraph-of-thought (HoT) reasoning paradigm, enabling the foundation models to think like an expert and possess the abilities of high-order multi-hop reasoning and multimodal comparative judgement. Specifically, based on the representation ability of the language foundation model, a textual HoT is constructed using triplets as the basic thought, and multi-hop logical reasoning is achieved by combining multi-hop walking paths. Furthermore, with the image features from a visual foundation model, a visual HoT is established, and cross-modal co-attention graph learning is chosen to interact with the textual HoT, realizing multimodal comparative judgement. The experimental results in the scienceQA dataset indicate that the proposed HoT-enhanced foundation models outperform CoT-based GPT3.5 and chatGPT.

Although the current multimodal HoT has not achieved as excellent performance as imagined, it can open up a new paradigm for instruction learning in foundation models. We believe multimodal HoT will tap its potential in the increasingly popular era of multimodal foundation models in the future. The HoT will be hot!

\bibliographystyle{unsrt}
\bibliography{references}

\begin{thebibliography}{10}

\bibitem{DBLP:journals/corr/abs-2304-09842}
Pan Lu, Baolin Peng, Hao Cheng, Michel Galley, Kai{-}Wei Chang, Ying~Nian Wu,
  Song{-}Chun Zhu, and Jianfeng Gao.
\newblock Chameleon: Plug-and-play compositional reasoning with large language
  models.
\newblock {\em CoRR}, abs/2304.09842, 2023.

\bibitem{yang2023harnessing}
Jingfeng Yang, Hongye Jin, Ruixiang Tang, Xiaotian Han, Qizhang Feng, Haoming
  Jiang, Bing Yin, and Xia Hu.
\newblock Harnessing the power of llms in practice: A survey on chatgpt and
  beyond, 2023.

\bibitem{chowdhery2022palm}
Aakanksha Chowdhery, Sharan Narang, Jacob Devlin, Maarten Bosma, Gaurav Mishra,
  Adam Roberts, Paul Barham, Hyung~Won Chung, Charles Sutton, Sebastian
  Gehrmann, Parker Schuh, Kensen Shi, Sasha Tsvyashchenko, Joshua Maynez,
  Abhishek Rao, Parker Barnes, Yi~Tay, Noam Shazeer, Vinodkumar Prabhakaran,
  Emily Reif, Nan Du, Ben Hutchinson, Reiner Pope, James Bradbury, Jacob
  Austin, Michael Isard, Guy Gur-Ari, Pengcheng Yin, Toju Duke, Anselm
  Levskaya, Sanjay Ghemawat, Sunipa Dev, Henryk Michalewski, Xavier Garcia,
  Vedant Misra, Kevin Robinson, Liam Fedus, Denny Zhou, Daphne Ippolito, David
  Luan, Hyeontaek Lim, Barret Zoph, Alexander Spiridonov, Ryan Sepassi, David
  Dohan, Shivani Agrawal, Mark Omernick, Andrew~M. Dai,
  Thanumalayan~Sankaranarayana Pillai, Marie Pellat, Aitor Lewkowycz, Erica
  Moreira, Rewon Child, Oleksandr Polozov, Katherine Lee, Zongwei Zhou, Xuezhi
  Wang, Brennan Saeta, Mark Diaz, Orhan Firat, Michele Catasta, Jason Wei,
  Kathy Meier-Hellstern, Douglas Eck, Jeff Dean, Slav Petrov, and Noah Fiedel.
\newblock Palm: Scaling language modeling with pathways, 2022.

\bibitem{zhang2022opt}
Susan Zhang, Stephen Roller, Naman Goyal, Mikel Artetxe, Moya Chen, Shuohui
  Chen, Christopher Dewan, Mona Diab, Xian Li, Xi~Victoria Lin, Todor Mihaylov,
  Myle Ott, Sam Shleifer, Kurt Shuster, Daniel Simig, Punit~Singh Koura, Anjali
  Sridhar, Tianlu Wang, and Luke Zettlemoyer.
\newblock Opt: Open pre-trained transformer language models, 2022.

\bibitem{workshop2023bloom}
BigScience Workshop, :, Teven~Le Scao, Angela Fan, Christopher Akiki, Ellie
  Pavlick, Suzana Ilić, Daniel Hesslow, Roman Castagné, Alexandra~Sasha
  Luccioni, François Yvon, Matthias Gallé, Jonathan Tow, Alexander~M. Rush,
  Stella Biderman, Albert Webson, Pawan~Sasanka Ammanamanchi, Thomas Wang,
  Benoît Sagot, Niklas Muennighoff, Albert~Villanova del Moral, Olatunji
  Ruwase, Rachel Bawden, Stas Bekman, Angelina McMillan-Major, Iz~Beltagy, Huu
  Nguyen, Lucile Saulnier, Samson Tan, Pedro~Ortiz Suarez, Victor Sanh, Hugo
  Laurençon, Yacine Jernite, Julien Launay, Margaret Mitchell, Colin Raffel,
  Aaron Gokaslan, Adi Simhi, Aitor Soroa, Alham~Fikri Aji, Amit Alfassy, Anna
  Rogers, Ariel~Kreisberg Nitzav, Canwen Xu, Chenghao Mou, Chris Emezue,
  Christopher Klamm, Colin Leong, Daniel van Strien, David~Ifeoluwa Adelani,
  Dragomir Radev, Eduardo~González Ponferrada, Efrat Levkovizh, Ethan Kim,
  Eyal~Bar Natan, Francesco~De Toni, Gérard Dupont, Germán Kruszewski, Giada
  Pistilli, Hady Elsahar, Hamza Benyamina, Hieu Tran, Ian Yu, Idris Abdulmumin,
  Isaac Johnson, Itziar Gonzalez-Dios, Javier de~la Rosa, Jenny Chim, Jesse
  Dodge, Jian Zhu, Jonathan Chang, Jörg Frohberg, Joseph Tobing, Joydeep
  Bhattacharjee, Khalid Almubarak, Kimbo Chen, Kyle Lo, Leandro~Von Werra, Leon
  Weber, Long Phan, Loubna~Ben allal, Ludovic Tanguy, Manan Dey, Manuel~Romero
  Muñoz, Maraim Masoud, María Grandury, Mario Šaško, Max Huang, Maximin
  Coavoux, Mayank Singh, Mike Tian-Jian Jiang, Minh~Chien Vu, Mohammad~A.
  Jauhar, Mustafa Ghaleb, Nishant Subramani, Nora Kassner, Nurulaqilla Khamis,
  Olivier Nguyen, Omar Espejel, Ona de~Gibert, Paulo Villegas, Peter Henderson,
  Pierre Colombo, Priscilla Amuok, Quentin Lhoest, Rheza Harliman, Rishi
  Bommasani, Roberto~Luis López, Rui Ribeiro, Salomey Osei, Sampo Pyysalo,
  Sebastian Nagel, Shamik Bose, Shamsuddeen~Hassan Muhammad, Shanya Sharma,
  Shayne Longpre, Somaieh Nikpoor, Stanislav Silberberg, Suhas Pai, Sydney
  Zink, Tiago~Timponi Torrent, Timo Schick, Tristan Thrush, Valentin Danchev,
  Vassilina Nikoulina, Veronika Laippala, Violette Lepercq, Vrinda Prabhu, Zaid
  Alyafeai, Zeerak Talat, Arun Raja, Benjamin Heinzerling, Chenglei Si,
  Davut~Emre Taşar, Elizabeth Salesky, Sabrina~J. Mielke, Wilson~Y. Lee,
  Abheesht Sharma, Andrea Santilli, Antoine Chaffin, Arnaud Stiegler, Debajyoti
  Datta, Eliza Szczechla, Gunjan Chhablani, Han Wang, Harshit Pandey, Hendrik
  Strobelt, Jason~Alan Fries, Jos Rozen, Leo Gao, Lintang Sutawika, M~Saiful
  Bari, Maged~S. Al-shaibani, Matteo Manica, Nihal Nayak, Ryan Teehan, Samuel
  Albanie, Sheng Shen, Srulik Ben-David, Stephen~H. Bach, Taewoon Kim, Tali
  Bers, Thibault Fevry, Trishala Neeraj, Urmish Thakker, Vikas Raunak, Xiangru
  Tang, Zheng-Xin Yong, Zhiqing Sun, Shaked Brody, Yallow Uri, Hadar Tojarieh,
  Adam Roberts, Hyung~Won Chung, Jaesung Tae, Jason Phang, Ofir Press, Conglong
  Li, Deepak Narayanan, Hatim Bourfoune, Jared Casper, Jeff Rasley, Max
  Ryabinin, Mayank Mishra, Minjia Zhang, Mohammad Shoeybi, Myriam Peyrounette,
  Nicolas Patry, Nouamane Tazi, Omar Sanseviero, Patrick von Platen, Pierre
  Cornette, Pierre~François Lavallée, Rémi Lacroix, Samyam Rajbhandari,
  Sanchit Gandhi, Shaden Smith, Stéphane Requena, Suraj Patil, Tim Dettmers,
  Ahmed Baruwa, Amanpreet Singh, Anastasia Cheveleva, Anne-Laure Ligozat, Arjun
  Subramonian, Aurélie Névéol, Charles Lovering, Dan Garrette, Deepak
  Tunuguntla, Ehud Reiter, Ekaterina Taktasheva, Ekaterina Voloshina, Eli
  Bogdanov, Genta~Indra Winata, Hailey Schoelkopf, Jan-Christoph Kalo,
  Jekaterina Novikova, Jessica~Zosa Forde, Jordan Clive, Jungo Kasai, Ken
  Kawamura, Liam Hazan, Marine Carpuat, Miruna Clinciu, Najoung Kim, Newton
  Cheng, Oleg Serikov, Omer Antverg, Oskar van~der Wal, Rui Zhang, Ruochen
  Zhang, Sebastian Gehrmann, Shachar Mirkin, Shani Pais, Tatiana Shavrina,
  Thomas Scialom, Tian Yun, Tomasz Limisiewicz, Verena Rieser, Vitaly Protasov,
  Vladislav Mikhailov, Yada Pruksachatkun, Yonatan Belinkov, Zachary Bamberger,
  Zdeněk Kasner, Alice Rueda, Amanda Pestana, Amir Feizpour, Ammar Khan, Amy
  Faranak, Ana Santos, Anthony Hevia, Antigona Unldreaj, Arash Aghagol, Arezoo
  Abdollahi, Aycha Tammour, Azadeh HajiHosseini, Bahareh Behroozi, Benjamin
  Ajibade, Bharat Saxena, Carlos~Muñoz Ferrandis, Daniel McDuff, Danish
  Contractor, David Lansky, Davis David, Douwe Kiela, Duong~A. Nguyen, Edward
  Tan, Emi Baylor, Ezinwanne Ozoani, Fatima Mirza, Frankline Ononiwu, Habib
  Rezanejad, Hessie Jones, Indrani Bhattacharya, Irene Solaiman, Irina Sedenko,
  Isar Nejadgholi, Jesse Passmore, Josh Seltzer, Julio~Bonis Sanz, Livia Dutra,
  Mairon Samagaio, Maraim Elbadri, Margot Mieskes, Marissa Gerchick, Martha
  Akinlolu, Michael McKenna, Mike Qiu, Muhammed Ghauri, Mykola Burynok, Nafis
  Abrar, Nazneen Rajani, Nour Elkott, Nour Fahmy, Olanrewaju Samuel, Ran An,
  Rasmus Kromann, Ryan Hao, Samira Alizadeh, Sarmad Shubber, Silas Wang, Sourav
  Roy, Sylvain Viguier, Thanh Le, Tobi Oyebade, Trieu Le, Yoyo Yang, Zach
  Nguyen, Abhinav~Ramesh Kashyap, Alfredo Palasciano, Alison Callahan, Anima
  Shukla, Antonio Miranda-Escalada, Ayush Singh, Benjamin Beilharz, Bo~Wang,
  Caio Brito, Chenxi Zhou, Chirag Jain, Chuxin Xu, Clémentine Fourrier,
  Daniel~León Periñán, Daniel Molano, Dian Yu, Enrique Manjavacas, Fabio
  Barth, Florian Fuhrimann, Gabriel Altay, Giyaseddin Bayrak, Gully Burns,
  Helena~U. Vrabec, Imane Bello, Ishani Dash, Jihyun Kang, John Giorgi, Jonas
  Golde, Jose~David Posada, Karthik~Rangasai Sivaraman, Lokesh Bulchandani,
  Lu~Liu, Luisa Shinzato, Madeleine~Hahn de~Bykhovetz, Maiko Takeuchi, Marc
  Pàmies, Maria~A Castillo, Marianna Nezhurina, Mario Sänger, Matthias
  Samwald, Michael Cullan, Michael Weinberg, Michiel~De Wolf, Mina Mihaljcic,
  Minna Liu, Moritz Freidank, Myungsun Kang, Natasha Seelam, Nathan Dahlberg,
  Nicholas~Michio Broad, Nikolaus Muellner, Pascale Fung, Patrick Haller, Ramya
  Chandrasekhar, Renata Eisenberg, Robert Martin, Rodrigo Canalli, Rosaline Su,
  Ruisi Su, Samuel Cahyawijaya, Samuele Garda, Shlok~S Deshmukh, Shubhanshu
  Mishra, Sid Kiblawi, Simon Ott, Sinee Sang-aroonsiri, Srishti Kumar, Stefan
  Schweter, Sushil Bharati, Tanmay Laud, Théo Gigant, Tomoya Kainuma, Wojciech
  Kusa, Yanis Labrak, Yash~Shailesh Bajaj, Yash Venkatraman, Yifan Xu, Yingxin
  Xu, Yu~Xu, Zhe Tan, Zhongli Xie, Zifan Ye, Mathilde Bras, Younes Belkada, and
  Thomas Wolf.
\newblock Bloom: A 176b-parameter open-access multilingual language model,
  2023.

\bibitem{du2022glam}
Nan Du, Yanping Huang, Andrew~M. Dai, Simon Tong, Dmitry Lepikhin, Yuanzhong
  Xu, Maxim Krikun, Yanqi Zhou, Adams~Wei Yu, Orhan Firat, Barret Zoph, Liam
  Fedus, Maarten Bosma, Zongwei Zhou, Tao Wang, Yu~Emma Wang, Kellie Webster,
  Marie Pellat, Kevin Robinson, Kathleen Meier-Hellstern, Toju Duke, Lucas
  Dixon, Kun Zhang, Quoc~V Le, Yonghui Wu, Zhifeng Chen, and Claire Cui.
\newblock Glam: Efficient scaling of language models with mixture-of-experts,
  2022.

\bibitem{rae2022scaling}
Jack~W. Rae, Sebastian Borgeaud, Trevor Cai, Katie Millican, Jordan Hoffmann,
  Francis Song, John Aslanides, Sarah Henderson, Roman Ring, Susannah Young,
  Eliza Rutherford, Tom Hennigan, Jacob Menick, Albin Cassirer, Richard Powell,
  George van~den Driessche, Lisa~Anne Hendricks, Maribeth Rauh, Po-Sen Huang,
  Amelia Glaese, Johannes Welbl, Sumanth Dathathri, Saffron Huang, Jonathan
  Uesato, John Mellor, Irina Higgins, Antonia Creswell, Nat McAleese, Amy Wu,
  Erich Elsen, Siddhant Jayakumar, Elena Buchatskaya, David Budden, Esme
  Sutherland, Karen Simonyan, Michela Paganini, Laurent Sifre, Lena Martens,
  Xiang~Lorraine Li, Adhiguna Kuncoro, Aida Nematzadeh, Elena Gribovskaya,
  Domenic Donato, Angeliki Lazaridou, Arthur Mensch, Jean-Baptiste Lespiau,
  Maria Tsimpoukelli, Nikolai Grigorev, Doug Fritz, Thibault Sottiaux, Mantas
  Pajarskas, Toby Pohlen, Zhitao Gong, Daniel Toyama, Cyprien
  de~Masson~d'Autume, Yujia Li, Tayfun Terzi, Vladimir Mikulik, Igor
  Babuschkin, Aidan Clark, Diego de~Las~Casas, Aurelia Guy, Chris Jones, James
  Bradbury, Matthew Johnson, Blake Hechtman, Laura Weidinger, Iason Gabriel,
  William Isaac, Ed~Lockhart, Simon Osindero, Laura Rimell, Chris Dyer, Oriol
  Vinyals, Kareem Ayoub, Jeff Stanway, Lorrayne Bennett, Demis Hassabis, Koray
  Kavukcuoglu, and Geoffrey Irving.
\newblock Scaling language models: Methods, analysis \& insights from training
  gopher, 2022.

\bibitem{hoffmann2022training}
Jordan Hoffmann, Sebastian Borgeaud, Arthur Mensch, Elena Buchatskaya, Trevor
  Cai, Eliza Rutherford, Diego de~Las~Casas, Lisa~Anne Hendricks, Johannes
  Welbl, Aidan Clark, Tom Hennigan, Eric Noland, Katie Millican, George van~den
  Driessche, Bogdan Damoc, Aurelia Guy, Simon Osindero, Karen Simonyan, Erich
  Elsen, Jack~W. Rae, Oriol Vinyals, and Laurent Sifre.
\newblock Training compute-optimal large language models, 2022.

\bibitem{thoppilan2022lamda}
Romal Thoppilan, Daniel~De Freitas, Jamie Hall, Noam Shazeer, Apoorv
  Kulshreshtha, Heng-Tze Cheng, Alicia Jin, Taylor Bos, Leslie Baker, Yu~Du,
  YaGuang Li, Hongrae Lee, Huaixiu~Steven Zheng, Amin Ghafouri, Marcelo
  Menegali, Yanping Huang, Maxim Krikun, Dmitry Lepikhin, James Qin, Dehao
  Chen, Yuanzhong Xu, Zhifeng Chen, Adam Roberts, Maarten Bosma, Vincent Zhao,
  Yanqi Zhou, Chung-Ching Chang, Igor Krivokon, Will Rusch, Marc Pickett,
  Pranesh Srinivasan, Laichee Man, Kathleen Meier-Hellstern, Meredith~Ringel
  Morris, Tulsee Doshi, Renelito~Delos Santos, Toju Duke, Johnny Soraker, Ben
  Zevenbergen, Vinodkumar Prabhakaran, Mark Diaz, Ben Hutchinson, Kristen
  Olson, Alejandra Molina, Erin Hoffman-John, Josh Lee, Lora Aroyo, Ravi
  Rajakumar, Alena Butryna, Matthew Lamm, Viktoriya Kuzmina, Joe Fenton, Aaron
  Cohen, Rachel Bernstein, Ray Kurzweil, Blaise Aguera-Arcas, Claire Cui,
  Marian Croak, Ed~Chi, and Quoc Le.
\newblock Lamda: Language models for dialog applications, 2022.

\bibitem{touvron2023llama}
Hugo Touvron, Thibaut Lavril, Gautier Izacard, Xavier Martinet, Marie-Anne
  Lachaux, Timothée Lacroix, Baptiste Rozière, Naman Goyal, Eric Hambro,
  Faisal Azhar, Aurelien Rodriguez, Armand Joulin, Edouard Grave, and Guillaume
  Lample.
\newblock Llama: Open and efficient foundation language models, 2023.

\bibitem{openai2023gpt4}
OpenAI.
\newblock Gpt-4 technical report, 2023.

\bibitem{wu2023bloomberggpt}
Shijie Wu, Ozan Irsoy, Steven Lu, Vadim Dabravolski, Mark Dredze, Sebastian
  Gehrmann, Prabhanjan Kambadur, David Rosenberg, and Gideon Mann.
\newblock Bloomberggpt: A large language model for finance, 2023.

\bibitem{lu2022learn}
Pan Lu, Swaroop Mishra, Tanglin Xia, Liang Qiu, Kai-Wei Chang, Song-Chun Zhu,
  Oyvind Tafjord, Peter Clark, and Ashwin Kalyan.
\newblock Learn to explain: Multimodal reasoning via thought chains for science
  question answering.
\newblock {\em Advances in Neural Information Processing Systems},
  35:2507--2521, 2022.

\bibitem{wei2023chainofthought}
Jason Wei, Xuezhi Wang, Dale Schuurmans, Maarten Bosma, Brian Ichter, Fei Xia,
  Ed~Chi, Quoc Le, and Denny Zhou.
\newblock Chain-of-thought prompting elicits reasoning in large language
  models, 2023.

\bibitem{wang2023selfconsistency}
Xuezhi Wang, Jason Wei, Dale Schuurmans, Quoc Le, Ed~Chi, Sharan Narang,
  Aakanksha Chowdhery, and Denny Zhou.
\newblock Self-consistency improves chain of thought reasoning in language
  models, 2023.

\bibitem{ning2023skeletonofthought}
Xuefei Ning, Zinan Lin, Zixuan Zhou, Huazhong Yang, and Yu~Wang.
\newblock Skeleton-of-thought: Large language models can do parallel decoding,
  2023.

\bibitem{yao2023tree}
Shunyu Yao, Dian Yu, Jeffrey Zhao, Izhak Shafran, Thomas~L. Griffiths, Yuan
  Cao, and Karthik Narasimhan.
\newblock Tree of thoughts: Deliberate problem solving with large language
  models, 2023.

\bibitem{yao2023chainofthought}
Yao Yao, Zuchao Li, and Hai Zhao.
\newblock Beyond chain-of-thought, effective graph-of-thought reasoning in
  large language models, 2023.

\bibitem{10068184}
Xian Sun, Fanglong Yao, and Chibiao Ding.
\newblock Modeling high-order relationships: Brain-inspired hypergraph-induced
  multimodal-multitask framework for semantic comprehension.
\newblock {\em IEEE Transactions on Neural Networks and Learning Systems},
  pages 1--15, 2023.

\bibitem{DBLP:journals/corr/abs-2302-00923}
Zhuosheng Zhang, Aston Zhang, Mu~Li, Hai Zhao, George Karypis, and Alex Smola.
\newblock Multimodal chain-of-thought reasoning in language models.
\newblock {\em CoRR}, abs/2302.00923, 2023.

\bibitem{kojima2023large}
Takeshi Kojima, Shixiang~Shane Gu, Machel Reid, Yutaka Matsuo, and Yusuke
  Iwasawa.
\newblock Large language models are zero-shot reasoners, 2023.

\bibitem{ho2022large}
Namgyu Ho, Laura Schmid, and Se-Young Yun.
\newblock Large language models are reasoning teachers.
\newblock {\em arXiv preprint arXiv:2212.10071}, 2022.

\bibitem{magister2022teaching}
Lucie~Charlotte Magister, Jonathan Mallinson, Jakub Adamek, Eric Malmi, and
  Aliaksei Severyn.
\newblock Teaching small language models to reason.
\newblock {\em arXiv preprint arXiv:2212.08410}, 2022.

\bibitem{li2023symbolic}
Liunian~Harold Li, Jack Hessel, Youngjae Yu, Xiang Ren, Kai-Wei Chang, and
  Yejin Choi.
\newblock Symbolic chain-of-thought distillation: Small models can also" think"
  step-by-step.
\newblock {\em arXiv preprint arXiv:2306.14050}, 2023.

\bibitem{zhang2023multimodal}
Zhuosheng Zhang, Aston Zhang, Mu~Li, Hai Zhao, George Karypis, and Alex Smola.
\newblock Multimodal chain-of-thought reasoning in language models.
\newblock {\em arXiv preprint arXiv:2302.00923}, 2023.

\bibitem{yao2023beyond}
Yao Yao, Zuchao Li, and Hai Zhao.
\newblock Beyond chain-of-thought, effective graph-of-thought reasoning in
  large language models.
\newblock {\em arXiv preprint arXiv:2305.16582}, 2023.

\bibitem{liu2023visual}
Haotian Liu, Chunyuan Li, Qingyang Wu, and Yong~Jae Lee.
\newblock Visual instruction tuning.
\newblock {\em arXiv preprint arXiv:2304.08485}, 2023.

\bibitem{luo2023cheap}
Gen Luo, Yiyi Zhou, Tianhe Ren, Shengxin Chen, Xiaoshuai Sun, and Rongrong Ji.
\newblock Cheap and quick: Efficient vision-language instruction tuning for
  large language models.
\newblock {\em arXiv preprint arXiv:2305.15023}, 2023.

\bibitem{wang2023t}
Lei Wang, Yi~Hu, Jiabang He, Xing Xu, Ning Liu, Hui Liu, and Heng~Tao Shen.
\newblock T-sciq: Teaching multimodal chain-of-thought reasoning via large
  language model signals for science question answering.
\newblock {\em arXiv preprint arXiv:2305.03453}, 2023.

\bibitem{horawalavithana2023scitune}
Sameera Horawalavithana, Sai Munikoti, Ian Stewart, and Henry Kvinge.
\newblock Scitune: Aligning large language models with scientific multimodal
  instructions.
\newblock {\em arXiv preprint arXiv:2307.01139}, 2023.

\bibitem{heo2022hypergraph}
Yu-Jung Heo, Eun-Sol Kim, Woo~Suk Choi, and Byoung-Tak Zhang.
\newblock Hypergraph transformer: Weakly-supervised multi-hop reasoning for
  knowledge-based visual question answering.
\newblock {\em arXiv preprint arXiv:2204.10448}, 2022.

\bibitem{kim2020hypergraph}
Eun-Sol Kim, Woo~Young Kang, Kyoung-Woon On, Yu-Jung Heo, and Byoung-Tak Zhang.
\newblock Hypergraph attention networks for multimodal learning.
\newblock In {\em Proceedings of the IEEE/CVF conference on computer vision and
  pattern recognition}, pages 14581--14590, 2020.

\bibitem{TIAN2022263}
Yu~Tian, Xian Sun, Ruigang Niu, Hongfeng Yu, Zicong Zhu, Peijin Wang, and Kun
  Fu.
\newblock Fully-weighted hgnn: Learning efficient non-local relations with
  hypergraph in aerial imagery.
\newblock {\em ISPRS Journal of Photogrammetry and Remote Sensing},
  191:263--276, 2022.

\bibitem{liu2021swin}
Ze~Liu, Yutong Lin, Yue Cao, Han Hu, Yixuan Wei, Zheng Zhang, Stephen Lin, and
  Baining Guo.
\newblock Swin transformer: Hierarchical vision transformer using shifted
  windows.
\newblock In {\em Proceedings of the IEEE/CVF international conference on
  computer vision}, pages 10012--10022, 2021.

\bibitem{gaohyper}
Yue Gao, Zizhao Zhang, Haojie Lin, Xibin Zhao, Shaoyi Du, and Changqing Zou.
\newblock Hypergraph learning: Methods and practices.
\newblock {\em IEEE Transactions on Pattern Analysis and Machine Intelligence},
  44(5):2548--2566, 2022.

\bibitem{raffel2020exploring}
Colin Raffel, Noam Shazeer, Adam Roberts, Katherine Lee, Sharan Narang, Michael
  Matena, Yanqi Zhou, Wei Li, and Peter~J Liu.
\newblock Exploring the limits of transfer learning with a unified text-to-text
  transformer.
\newblock {\em The Journal of Machine Learning Research}, 21(1):5485--5551,
  2020.

\bibitem{DBLP:conf/emnlp/KhashabiMKSTCH20}
Daniel Khashabi, Sewon Min, Tushar Khot, Ashish Sabharwal, Oyvind Tafjord,
  Peter Clark, and Hannaneh Hajishirzi.
\newblock Unifiedqa: Crossing format boundaries with a single {QA} system.
\newblock In Trevor Cohn, Yulan He, and Yang Liu, editors, {\em Findings of the
  Association for Computational Linguistics: {EMNLP} 2020, Online Event, 16-20
  November 2020}, volume {EMNLP} 2020 of {\em Findings of {ACL}}, pages
  1896--1907. Association for Computational Linguistics, 2020.

\bibitem{DBLP:conf/eccv/CarionMSUKZ20}
Nicolas Carion, Francisco Massa, Gabriel Synnaeve, Nicolas Usunier, Alexander
  Kirillov, and Sergey Zagoruyko.
\newblock End-to-end object detection with transformers.
\newblock In Andrea Vedaldi, Horst Bischof, Thomas Brox, and Jan{-}Michael
  Frahm, editors, {\em Computer Vision - {ECCV} 2020 - 16th European
  Conference, Glasgow, UK, August 23-28, 2020, Proceedings, Part {I}}, volume
  12346 of {\em Lecture Notes in Computer Science}, pages 213--229. Springer,
  2020.

\bibitem{DBLP:conf/cvpr/Yu0CT019}
Zhou Yu, Jun Yu, Yuhao Cui, Dacheng Tao, and Qi~Tian.
\newblock Deep modular co-attention networks for visual question answering.
\newblock In {\em {IEEE} Conference on Computer Vision and Pattern Recognition,
  {CVPR} 2019, Long Beach, CA, USA, June 16-20, 2019}, pages 6281--6290.
  Computer Vision Foundation / {IEEE}, 2019.

\bibitem{DBLP:conf/cvpr/00010BT0GZ18}
Peter Anderson, Xiaodong He, Chris Buehler, Damien Teney, Mark Johnson, Stephen
  Gould, and Lei Zhang.
\newblock Bottom-up and top-down attention for image captioning and visual
  question answering.
\newblock In {\em 2018 {IEEE} Conference on Computer Vision and Pattern
  Recognition, {CVPR} 2018, Salt Lake City, UT, USA, June 18-22, 2018}, pages
  6077--6086. Computer Vision Foundation / {IEEE} Computer Society, 2018.

\bibitem{DBLP:conf/nips/KimJZ18}
Jin{-}Hwa Kim, Jaehyun Jun, and Byoung{-}Tak Zhang.
\newblock Bilinear attention networks.
\newblock In Samy Bengio, Hanna~M. Wallach, Hugo Larochelle, Kristen Grauman,
  Nicol{\`{o}} Cesa{-}Bianchi, and Roman Garnett, editors, {\em Advances in
  Neural Information Processing Systems 31: Annual Conference on Neural
  Information Processing Systems 2018, NeurIPS 2018, December 3-8, 2018,
  Montr{\'{e}}al, Canada}, pages 1571--1581, 2018.

\bibitem{DBLP:conf/cvpr/GaoJYLHWL19}
Peng Gao, Zhengkai Jiang, Haoxuan You, Pan Lu, Steven C.~H. Hoi, Xiaogang Wang,
  and Hongsheng Li.
\newblock Dynamic fusion with intra- and inter-modality attention flow for
  visual question answering.
\newblock In {\em {IEEE} Conference on Computer Vision and Pattern Recognition,
  {CVPR} 2019, Long Beach, CA, USA, June 16-20, 2019}, pages 6639--6648.
  Computer Vision Foundation / {IEEE}, 2019.

\bibitem{DBLP:conf/nips/LuQCXZZYLZ21}
Pan Lu, Liang Qiu, Jiaqi Chen, Tanglin Xia, Yizhou Zhao, Wei Zhang, Zhou Yu,
  Xiaodan Liang, and Song{-}Chun Zhu.
\newblock Iconqa: {A} new benchmark for abstract diagram understanding and
  visual language reasoning.
\newblock In Joaquin Vanschoren and Sai{-}Kit Yeung, editors, {\em Proceedings
  of the Neural Information Processing Systems Track on Datasets and Benchmarks
  1, NeurIPS Datasets and Benchmarks 2021, December 2021, virtual}, 2021.

\bibitem{DBLP:journals/corr/abs-1908-03557}
Liunian~Harold Li, Mark Yatskar, Da~Yin, Cho{-}Jui Hsieh, and Kai{-}Wei Chang.
\newblock Visualbert: {A} simple and performant baseline for vision and
  language.
\newblock {\em CoRR}, abs/1908.03557, 2019.

\bibitem{DBLP:conf/icml/KimSK21}
Wonjae Kim, Bokyung Son, and Ildoo Kim.
\newblock Vilt: Vision-and-language transformer without convolution or region
  supervision.
\newblock In Marina Meila and Tong Zhang, editors, {\em Proceedings of the 38th
  International Conference on Machine Learning, {ICML} 2021, 18-24 July 2021,
  Virtual Event}, volume 139 of {\em Proceedings of Machine Learning Research},
  pages 5583--5594. {PMLR}, 2021.

\bibitem{DBLP:conf/nips/ChenKSNH20}
Ting Chen, Simon Kornblith, Kevin Swersky, Mohammad Norouzi, and Geoffrey~E.
  Hinton.
\newblock Big self-supervised models are strong semi-supervised learners.
\newblock In Hugo Larochelle, Marc'Aurelio Ranzato, Raia Hadsell,
  Maria{-}Florina Balcan, and Hsuan{-}Tien Lin, editors, {\em Advances in
  Neural Information Processing Systems 33: Annual Conference on Neural
  Information Processing Systems 2020, NeurIPS 2020, December 6-12, 2020,
  virtual}, 2020.

\end{thebibliography}

\end{document}